\documentclass[10pt,twocolumn,letterpaper]{article}

\usepackage[pagenumbers]{cvpr} 

\usepackage[dvipsnames]{xcolor}
\definecolor{cvprblue}{rgb}{0.21,0.49,0.74}
\usepackage[pagebackref,breaklinks,colorlinks,linkcolor=cvprblue]{hyperref}

\usepackage{caption}
\usepackage{subcaption}
\usepackage{siunitx}
\usepackage{nicefrac, xfrac}

\usepackage{adjustbox}
\usepackage{array}
\usepackage{dblfloatfix}
\usepackage[english]{babel}

\addto\extrasenglish{

}

\newcolumntype{R}[2]{%
	>{\adjustbox{angle=#1,lap=\width-(#2)}\bgroup}%
	l%
	<{\egroup}%
}
\newcommand*\rot{\multicolumn{1}{R{30}{1em}}}

\def\sotaImprovement{64\%}
\def\sotaProgressImprovement{25\%}

\def\paperID{*****} 
\def\confName{CVPR}
\def\confYear{2024}

\title{Scaling Is All You Need: Autonomous Driving with JAX-Accelerated Reinforcement Learning}

\author{Moritz Harmel\\
{\tt\small zoox.com}
\and
Anubhav Paras \\
{\tt\small zoox.com}
\and
Andreas Pasternak \\
{\tt\small zoox.com}
\and
Nicholas Roy \\
{\tt\small zoox.com}
\and
Gary Linscott\\
{\tt\small zoox.com}
}

\begin{document}
\maketitle
\begin{abstract}
Reinforcement learning has been demonstrated to outperform even the best humans in complex domains like video games. 
However, running reinforcement learning experiments on the required scale for autonomous driving is extremely difficult. Building a large scale reinforcement learning system and distributing it across many GPUs is challenging. Gathering experience during training on real world vehicles is prohibitive from a safety and scalability perspective. Therefore, an efficient and realistic driving simulator is required that uses a large amount of data from real-world driving. 
We bring these capabilities together and conduct large-scale reinforcement learning experiments for autonomous driving. We demonstrate that our policy performance improves with increasing scale. Our best performing policy reduces the failure rate by  \sotaImprovement \,while improving the rate of driving progress by \sotaProgressImprovement \,compared to the policies produced by state-of-the-art machine learning for autonomous driving.
\end{abstract}
\vspace{-0.2cm}
\section{Introduction}

In this paper, we set up a scalable reinforcement learning framework and combine it with an efficient simulator for autonomous driving based on real-world data. We then conduct experiments on billions of agent steps with different model sizes in order to determine if we can overcome the constrained state space of the simulator by using increasingly more real-world data. 

The main contributions of our work are:
\begin{enumerate}
	\item We demonstrate how to use prerecorded real-world driving data in a hardware-accelerated simulator as part of distributed reinforcement learning to achieve improving policy performance with increasing experiment size.
	\item We demonstrate that our largest scale experiments, using a 25M parameter model on 6000 h of human-expert driving from San Francisco training on 2.5 billion agent steps, reduced the failure rate compared to the current state of the art \cite{lu2023imitationNotEnough} by \sotaImprovement.
\end{enumerate}


 \section{Related work}
 	\label{sec:related_work}
 	In this paper we present a hardware-accelerated autonomous driving simulator for real-world driving scenarios, which is similar to Waymax \cite{gulino2023waymax}. Waymax also trains RL agents but only reported results from 30 million agent steps, approximately two orders of magnitude less than our experiments.

    Training highly performant policies in complex and continuous real-world domains has mainly been achieved with distributed and scalable reinforcement learning using actor-critic methods. However, most existing results focus not on real-world problems but on video game environments \cite{vinyals2019alphastar, openai2019dota}. 

 	Of the techniques that have been used to train autonomous driving policies, the closest to our work is Imitation Is Not Enough \cite{lu2023imitationNotEnough} which also uses a combination of imitation learning (IL) and RL to train strong driving policies on a large dataset of real-world driving scenarios. That work established two fundamental driving metrics (failure rate and progress ratio) and provides a detailed description of the mining process of their evaluation dataset. The presented policies are state of the art (SOTA) and we will compare our best policy to theirs.
 	Other work focuses on realistic traffic simulation \cite{zhang2023learning}, but also trains policies with a combined IL and RL approach. However, their datasets are approximately 3 orders of magnitude smaller than ours and focus on highway driving, which does not pose as complex challenges as the dense urban driving we focus on.  Imitation learning approaches in open loop \cite{bronstein2022embedding, zhang2021endtoend, vitelli2021safetynet} and closed loop  \cite{bronstein2022hierarchical,  igl2022symphony} have also been proposed for autonomous driving. As \cite{lu2023imitationNotEnough} argue, IL approaches lack explicit knowledge of unsafe driving and can respond inappropriate in rare, long tail scenarios.
 
 	\section{Large scale RL for autonomous driving}
 	The challenges of using large scale reinforcement learning for autonomous driving are manifold.
 	To achieve scale, the real-world problem must be modeled by a simulator, which requires creating realistic traffic scenes and modeling the interactions between different dynamic agents. The simulator must be sufficiently efficient to generate environment interactions on the order of billions of steps in reasonable time and computational cost. Finally, a distributed learning architecture must be identified that can learn efficiently and scalably. Learners and simulation actors must be created across many machines, each leveraging parallel hardware, i.e. GPUs.
 	
 	\subsection{Scene generation and agent interactions}
 	\label{sec:scene_generation}
 	There are different ways of creating traffic scenes for simulation. Simulations scenarios can be generated entirely synthetically, including the placement  of roads, agents, and traffic signals, for example as in the Carla simulator \cite{dosovitskiy2017carla}. While entirely synthetic simulation gives very fine grained control over training scenario distribution, this approach raises the challenge of identifying what that scenario distribution should be, and how best to sample training instances from it. 
 	A different approach is to create scenes based on real-world data recorded from vehicles equipped with sensors, such as cameras and lidar sensors, driving on public roads. From the recorded sensor data, a 3D-scene can be created that contains, for example, the observed traffic light states and challenging obstacles, such as pedestrians, cyclists, and cars \cite{gulino2023waymax}. In this work, we use scenarios that have been created from real-world driving, for the fidelity of their representation of the real world.

 	\subsection{Accelerated autonomous driving simulator}
 	Because our ultimate goal is to run reinforcement learning experiments with billions of agent steps, the simulator must be very efficient. This can be achieved by running the simulation on accelerated hardware, such as GPUs. The parallel computing power of accelerated hardware can lead to massive speed ups compared to CPUs. However, challenges regarding the simulation data structures and control flow need to be addressed to enable large scale parallelism. In particular, all data need to be of the same size and no logical branches depending on the values of the data can be introduced. 

 	\subsubsection{Preparing data for parallel execution}
 	A traffic scene can be described by data that changes over time (dynamic data) and data that is constant throughout the scenario (static data). An example of dynamic data are the agents in the scene --- the number of agents can change during a scenario as well as between scenarios. An example for static data are the road segments, which do not change within a scenario but change across scenarios as different locations require different roads. Furthermore, the number of time steps per scene can vary.
 	
 	Data segments of different sizes are not suitable for parallel execution on hardware accelerators. To overcome this issue, a common maximum size for each type of data is defined. All data elements in the dynamic data are then padded to the defined maximum size for each time step. 

 	\subsubsection{The accelerated simulation utilizing JAX}
 	Conceptually, the simulation can be divided into a phase of action generation and a phase of advancing the environment state by applying the selected actions to the active agent and updating all other agent positions based on the logged trajectories. From the updated environment state including the recorded data, the required observations can be retrieved and used in the next step of action generation from the learning system as described in \autoref{sec:distributed_training}. The action generation is well-suited for batched inference, so the primary challenge in simulating on parallel hardware is to implement the environment update function to process batched data in parallel. We used the JAX library to rewrite the update function, so it can be jit-compiled and executed on batches on the GPU. Finally, to maximize the simulation speed, we combine the batched environment call with the batched model call and scan along the time axis of the dynamic data via the \textit{jax.lax.scan} primitive.  The entire simulation is then jit-compiled into a single graph and run in XLA.

    \subsubsection{Simulator performance benchmark}
	\label{sec:sim_benchmark}
	To demonstrate the performance of our simulator, we compare the environment step time with Waymax \cite{gulino2023waymax}, which is closest to our work. \autoref{tab:sim_runtime} reports the step time for different batch sizes on Nvidia v100 GPUs, showing that our simulator runs slightly faster. 

    \begin{table}[b]
	\caption{Runtime comparison for different batch sizes (BS) between Waymax \cite{gulino2023waymax} and our simulator for one controlled agent.} 
	\label{tab:sim_runtime}
	\centering
	\begin{tabular}{ll|lll}
		&   & \multicolumn{3}{c}{Step time [ms]}   \\
		Simulator & Device & BS  1 & BS 16 \\
		\midrule
		Waymax  & v100& 0.75  & 2.48 \\
		Ours          & v100 & 0.52 & 0.82  \\
		\bottomrule
	\end{tabular}
    \end{table}

 	\subsection{RL problem formulation}
 	For our reinforcement learning approach we need to specify how we retrieve the model inputs (observations) from the state $s$ and how we generate the physical actions from the model outputs. The state $s$ is the state of the simulation described previously, i.e., agents, roads, traffic lights etc. 
 	We also need to define the rewards, so the simulated data can be used to calculate the parameter update from an RL method.
	
 	\subsubsection{Observation space}
	\label{sec:observation_space}
	The observation space retrieves a subset of the information of the environment state and transforms the fields of the observation vector into an agent-centric coordinate frame.

	\subsubsection{Action space}
	Our model directly controls the longitudinal acceleration as well as the steering angle rate. Using these controls guarantees that the associated dynamic constraints are not violated. 
	We are using discrete actions, which we found to be more stable than continuous actions during reinforcement learning training.

	\subsubsection{Rewards}
	\label{sec:rewards}
	
	The goal of the policy is to navigate safely through traffic. In particular, the agent should make progress along the desired route while not colliding with other agents and adhering to basic traffic rules. In order to achieve  this, we introduce dense rewards as well as done signals that are associated with sparse rewards.
	
	Done signals have been introduced for collisions, off-route driving, as well as running red lights and stop lines and are associated with high negative rewards. Dense rewards are introduced for the progress along the planned route (positive), velocity above the speed limit, as well as on the squared lateral and longitudinal acceleration (all negative).
	
 	\subsection{Distributed learning system}
 	\label{sec:distributed_training}
 	
We can now establish the reinforcement learning approach. Our base reinforcement learner uses actor-critic Proximal Policy Optimization (PPO) \cite{schulman2017proximal}. However, to achieve scale, we set up an asynchronous reinforcement learning system similar to Dota2 \cite{openai2019dota}. The asynchronous setting allows to run the learners and actors independently avoiding any slow down. This in turn causes the data to be off-policy. We address this challenge for actor-critic methods by using the V-trace off-policy correction algorithm \cite{espeholt2018impala}. Furthermore, we pre-train a policy via behavioral cloning, similar to AlphaStar \cite{vinyals2019alphastar} because a good initial policy can speed up the RL training. However, only pre-training the policy and not the value network poses challenges to the stability at the start of RL training. Therefore, we use the discounted return of the expert trajectories as the value target. We calculate the discounted return by replaying the expert trajectories in our simulator and assigning the defined RL rewards.

	\section{Evaluation}
	\label{sec:evaluation}
	This section describes the different metrics and the dataset used for policy evaluation and comparison.
	\subsection{Metrics}
	\label{sec:metrics}
	The goal of the metrics are to measure the quality of the trained policy. This is already a complex problem for autonomous driving as the quality of driving comprises many different aspects. For the scope of this paper, we follow the work of \cite{lu2023imitationNotEnough} who introduced the failure rate and progress ratio as relevant metrics for autonomous driving. The failure rate measures the fundamental safety of the policy. If the agent collides or drives off-road in the simulation the scenario is considered as failed. 
    The progress ratio is the distance traveled by the agent in the simulation divided by the distance traveled of the vehicle in the original log. When the agent travels the same distance as the vehicle in the log, this metric becomes 100 \%.
	
	In addition to collisions and off-route failures, we also implemented metrics for stop line and traffic light violations. We did not include these violations in the failure rate to maintain consistency with previously reported results. However, we do report these metrics in \autoref{tab:main_experiment}.

	\subsection{Dataset}
	As the current state-of-the-art policies \cite{lu2023imitationNotEnough} are evaluated on proprietary datasets, our goal was to achieve the fairest comparison by following the same dataset mining procedure. A dataset of 10k randomly sampled 10 s segments was created using data collected from human-expert driving in San Francisco. This is comparable to the "All" evaluation dataset in \cite{lu2023imitationNotEnough}.

	\section{Experiments}
	\label{sec:experiments}
	
	Combining the real-world driving simulator, the scalable reinforcement learning framework and the described evaluation metrics and dataset, we conduct experiments with different training dataset and model sizes. For all these experiments we keep the hyperparameters the same. In particular we run our experiments for larger models across more GPUs to achieve the same batch size. 
	
	\begin{figure*}[htb]
		\centering
		\begin{subfigure}[T]{0.4\textwidth}
			\centering
			\includegraphics[trim={2.0cm 3.5cm 2.0cm 2.5cm},clip,width=\textwidth]{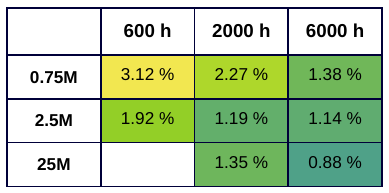}
			\caption{Minimum failure rate}
			\label{fig:failure_rate_table}
		\end{subfigure}
		\qquad
		\begin{subfigure}[T]{0.4\textwidth}
			\centering
			\includegraphics[trim={2.0cm 3.5cm 2.0cm 2.5cm},clip, width=\textwidth]{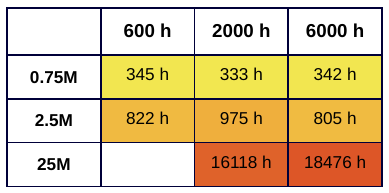}
			\caption{GPU time}
			\label{fig:gpu_time_table}
		\end{subfigure}
		\caption{Results for experiments with different model sizes (rows) and dataset sizes (columns). Colors represent the numerical results on color scales. (a) The performance of the policy improves with increasing model and dataset size. (b) The model size is the major driver of the required GPU time and therefore cost of training. Dataset size has no effect on the training time, but it can affect one time costs during data preprocessing which is not considered here. }
		\label{fig:scalability_experiments}
	\end{figure*}
	{
	\begin{table*}[b]
	\caption{Comparison of our policies with the current SOTA \cite{lu2023imitationNotEnough}.}
	\label{tab:main_experiment}
	\centering
	\begin{tabular}[H]{l|lllllll|}
		&
		\rot{BC \cite{lu2023imitationNotEnough}} &
		\rot{MGAIL \cite{lu2023imitationNotEnough}} &
		\rot{SAC \cite{lu2023imitationNotEnough}} &
		\rot{BC+SAC \cite{lu2023imitationNotEnough}} &
		\rot{Our BC} &
		\rot{Our BC+PPO} \\
		\midrule
		Failure Rate [\%]  \cite{lu2023imitationNotEnough} & 3.64 & 2.45 & 5.60 & 2.81  & 19.85 & \textbf{0.88}\\
		Progress Ratio [\%]  \cite{lu2023imitationNotEnough} & 98.1 & 96.6 & 71.1 & 87.6 & 96.91 & \textbf{120.8} \\
		\midrule
		Collisions [\%] & - & - & - &-& 10.32 &\textbf{0.46} \\
		Off-Route Events [\%] & - & - & - &- & 10.35 & \textbf{0.49} \\
		Stop Line Violations [\%] & - & - & - & - & 2.47 &\textbf{0.02} \\
		Traffic Light Violations [\%] & - & - & - & - & 2.08 & \textbf{0.28} \\
	\end{tabular}
\end{table*}
	}

	We mined three different training datasets of 600 h, 2000 h  and 6000 h from human-expert driving in San Francisco. We also created three different model sizes of 0.75M, 2.5M and 25M parameters by increasing the attention dimensions of the network. Each model is trained first by behavior cloning for 20 epochs on the given dataset. The pre-trained policy is then refined by reinforcement learning on 2.5B agent steps. We evaluate the policy during reinforcement learning every 20M agent steps and after training select the checkpoint with the lowest failure rate on the evaluation dataset.
	
	We conduct experiments on all combinations of model size and dataset size, with the exception of the small 600 h dataset in combination with the large 25M parameter model. \autoref{fig:failure_rate_table} shows that increasing the dataset size improves the performance of the trained policy in terms of failure rate. Increasing the model size in general also improves the policy performance. The 2.5M model is strictly better than the 0.75M model and the best policy is trained on the 25M model. However, we observe that increasing the model size only helps when sufficient real-world driving data is available. On the 2000 h dataset the 25M performs worse than the 2.5M model and only on the 6000 h dataset it performs better. The largest experiment achieves a failure rate of \SI{0.88}{\percent}.

	In \autoref{tab:main_experiment} we compare the policy performance of our largest setting after behavioral cloning and after reinforcement learning training with the current SOTA \cite{lu2023imitationNotEnough}. Our behavioral cloning policy performs quite poorly, achieving a failure rate of 19.85 \%. This is much higher than the pure BC failure rate of 3.64 \% reported in the current SOTA \cite{lu2023imitationNotEnough}. 
	
	The reinforcement learning training improves the policy and achieves a failure rate of 0.88 \% and a progress ratio of 120.8 \%. Compared to the best policy of the current SOTA on a similar dataset, the failure rate is reduced by \sotaImprovement \,and the progress ratio improved by \sotaProgressImprovement. 

	\section{Conclusions}
	\label{sec:conclusion}
	
	In this paper we combined an efficient and realistic autonomous driving simulator with a scalable reinforcement learning framework. This allowed us to run large scale reinforcement learning experiments training on billions of agents steps with increasing model size on different dataset sizes of real-world driving.

	Our data shows that we can obtain similar scaling behavior as in other reinforcement learning settings \cite{hilton2023scaling} when using increasingly large datasets of real-world driving. In particular, we were able to obtain better policies with larger models when using sufficiently large datasets. Our best policy reduces the failure rate compared to the current SOTA \cite{lu2023imitationNotEnough} by \sotaImprovement \,while improving progress by \sotaProgressImprovement. These results are very encouraging, and motivate further experiments with increasing size. However, to ultimately answer whether the presented approach can be scaled beyond human performance a validation framework that can reliably compare the safety of the policy to human drivers is also required.
	
{
    \small
    \bibliographystyle{ieeenat_fullname}
    \bibliography{zrl}

\begin{thebibliography}{14}
\providecommand{\natexlab}[1]{#1}
\providecommand{\url}[1]{\texttt{#1}}
\expandafter\ifx\csname urlstyle\endcsname\relax
  \providecommand{\doi}[1]{doi: #1}\else
  \providecommand{\doi}{doi: \begingroup \urlstyle{rm}\Url}\fi

\bibitem[Berner et~al.(2019)Berner, Brockman, Chan, Cheung, Debiak, Dennison,
  Farhi, Fischer, Hashme, Hesse, Józefowicz, Gray, Olsson, Pachocki, Petrov,
  d.~O.~Pinto, Raiman, Salimans, Schlatter, Schneider, Sidor, Sutskever, Tang,
  Wolski, and Zhang]{openai2019dota}
Christopher Berner, Greg Brockman, Brooke Chan, Vicki Cheung, Przemyslaw
  Debiak, Christy Dennison, David Farhi, Quirin Fischer, Shariq Hashme, Chris
  Hesse, Rafal Józefowicz, Scott Gray, Catherine Olsson, Jakub Pachocki,
  Michael Petrov, Henrique~P. d. O.~Pinto, Jonathan Raiman, Tim Salimans,
  Jeremy Schlatter, Jonas Schneider, Szymon Sidor, Ilya Sutskever, Jie Tang,
  Filip Wolski, and Susan Zhang.
\newblock Dota 2 with large scale deep reinforcement learning, 2019.

\bibitem[Bronstein et~al.(2022{\natexlab{a}})Bronstein, Palatucci, Notz, White,
  Kuefler, Lu, Paul, Nikdel, Mougin, Chen, Fu, Abrams, Shah, Racah, Frenkel,
  Whiteson, and Anguelov]{bronstein2022hierarchical}
Eli Bronstein, Mark Palatucci, Dominik Notz, Brandyn White, Alex Kuefler, Yiren
  Lu, Supratik Paul, Payam Nikdel, Paul Mougin, Hongge Chen, Justin Fu, Austin
  Abrams, Punit Shah, Evan Racah, Benjamin Frenkel, Shimon Whiteson, and
  Dragomir Anguelov.
\newblock Hierarchical model-based imitation learning for planning in
  autonomous driving, 2022{\natexlab{a}}.

\bibitem[Bronstein et~al.(2022{\natexlab{b}})Bronstein, Srinivasan, Paul,
  Sinha, O'Kelly, Nikdel, and Whiteson]{bronstein2022embedding}
Eli Bronstein, Sirish Srinivasan, Supratik Paul, Aman Sinha, Matthew O'Kelly,
  Payam Nikdel, and Shimon Whiteson.
\newblock Embedding synthetic off-policy experience for autonomous driving via
  zero-shot curricula, 2022{\natexlab{b}}.

\bibitem[Dosovitskiy et~al.(2017)Dosovitskiy, Ros, Codevilla, Lopez, and
  Koltun]{dosovitskiy2017carla}
Alexey Dosovitskiy, German Ros, Felipe Codevilla, Antonio Lopez, and Vladlen
  Koltun.
\newblock Carla: An open urban driving simulator, 2017.

\bibitem[Espeholt et~al.(2018)Espeholt, Soyer, Munos, Simonyan, Mnih, Ward,
  Doron, Firoiu, Harley, Dunning, Legg, and Kavukcuoglu]{espeholt2018impala}
Lasse Espeholt, Hubert Soyer, Remi Munos, Karen Simonyan, Volodymir Mnih, Tom
  Ward, Yotam Doron, Vlad Firoiu, Tim Harley, Iain Dunning, Shane Legg, and
  Koray Kavukcuoglu.
\newblock Impala: Scalable distributed deep-rl with importance weighted
  actor-learner architectures, 2018.

\bibitem[Gulino et~al.(2023)Gulino, Fu, Luo, Tucker, Bronstein, Lu, Harb, Pan,
  Wang, Chen, Co-Reyes, Agarwal, Roelofs, Lu, Montali, Mougin, Yang, White,
  Faust, McAllister, Anguelov, and Sapp]{gulino2023waymax}
Cole Gulino, Justin Fu, Wenjie Luo, George Tucker, Eli Bronstein, Yiren Lu,
  Jean Harb, Xinlei Pan, Yan Wang, Xiangyu Chen, John~D. Co-Reyes, Rishabh
  Agarwal, Rebecca Roelofs, Yao Lu, Nico Montali, Paul Mougin, Zoey Yang,
  Brandyn White, Aleksandra Faust, Rowan McAllister, Dragomir Anguelov, and
  Benjamin Sapp.
\newblock Waymax: An accelerated, data-driven simulator for large-scale
  autonomous driving research, 2023.

\bibitem[Hilton et~al.(2023)Hilton, Tang, and Schulman]{hilton2023scaling}
Jacob Hilton, Jie Tang, and John Schulman.
\newblock Scaling laws for single-agent reinforcement learning, 2023.

\bibitem[Igl et~al.(2022)Igl, Kim, Kuefler, Mougin, Shah, Shiarlis, Anguelov,
  Palatucci, White, and Whiteson]{igl2022symphony}
Maximilian Igl, Daewoo Kim, Alex Kuefler, Paul Mougin, Punit Shah, Kyriacos
  Shiarlis, Dragomir Anguelov, Mark Palatucci, Brandyn White, and Shimon
  Whiteson.
\newblock Symphony: Learning realistic and diverse agents for autonomous
  driving simulation, 2022.

\bibitem[Lu et~al.(2023)Lu, Fu, Tucker, Pan, Bronstein, Roelofs, Sapp, White,
  Faust, Whiteson, Anguelov, and Levine]{lu2023imitationNotEnough}
Yiren Lu, Justin Fu, George Tucker, Xinlei Pan, Eli Bronstein, Rebecca Roelofs,
  Benjamin Sapp, Brandyn White, Aleksandra Faust, Shimon Whiteson, Dragomir
  Anguelov, and Sergey Levine.
\newblock Imitation is not enough: Robustifying imitation with reinforcement
  learning for challenging driving scenarios, 2023.

\bibitem[Schulman et~al.(2017)Schulman, Wolski, Dhariwal, Radford, and
  Klimov]{schulman2017proximal}
John Schulman, Filip Wolski, Prafulla Dhariwal, Alec Radford, and Oleg Klimov.
\newblock Proximal policy optimization algorithms, 2017.

\bibitem[Vinyals et~al.(2019)Vinyals, Babuschkin, Czarnecki, Mathieu, Dudzik,
  Chung, Choi, Powell, Ewalds, Georgiev, Oh, Horgan, Kroiss, Danihelka, Huang,
  Sifre, Cai, Agapiou, Jaderberg, Vezhnevets, Leblond, Pohlen, Dalibard,
  Budden, Sulsky, Molloy, Paine, Gulcehre, Wang, Pfaff, Wu, Ring, Yogatama,
  W{\"u}nsch, McKinney, Smith, Schaul, Lillicrap, Kavukcuoglu, Hassabis, Apps,
  and Silver]{vinyals2019alphastar}
Oriol Vinyals, Igor Babuschkin, Wojciech~M. Czarnecki, Micha{\"e}l Mathieu,
  Andrew Dudzik, Junyoung Chung, David~H. Choi, Richard Powell, Timo Ewalds,
  Petko Georgiev, Junhyuk Oh, Dan Horgan, Manuel Kroiss, Ivo Danihelka, Aja
  Huang, Laurent Sifre, Trevor Cai, John~P. Agapiou, Max Jaderberg,
  Alexander~S. Vezhnevets, R{\'e}mi Leblond, Tobias Pohlen, Valentin Dalibard,
  David Budden, Yury Sulsky, James Molloy, Tom~L. Paine, Caglar Gulcehre, Ziyu
  Wang, Tobias Pfaff, Yuhuai Wu, Roman Ring, Dani Yogatama, Dario W{\"u}nsch,
  Katrina McKinney, Oliver Smith, Tom Schaul, Timothy Lillicrap, Koray
  Kavukcuoglu, Demis Hassabis, Chris Apps, and David Silver.
\newblock Grandmaster level in starcraft ii using multi-agent reinforcement
  learning.
\newblock \emph{Nature}, 575\penalty0 (7782):\penalty0 350--354, 2019.

\bibitem[Vitelli et~al.(2021)Vitelli, Chang, Ye, Wołczyk, Osiński, Niendorf,
  Grimmett, Huang, Jain, and Ondruska]{vitelli2021safetynet}
Matt Vitelli, Yan Chang, Yawei Ye, Maciej Wołczyk, Błażej Osiński, Moritz
  Niendorf, Hugo Grimmett, Qiangui Huang, Ashesh Jain, and Peter Ondruska.
\newblock Safetynet: Safe planning for real-world self-driving vehicles using
  machine-learned policies, 2021.

\bibitem[Zhang et~al.(2023)Zhang, Tu, Zhang, Wong, Suo, and
  Urtasun]{zhang2023learning}
Chris Zhang, James Tu, Lunjun Zhang, Kelvin Wong, Simon Suo, and Raquel
  Urtasun.
\newblock Learning realistic traffic agents in closed-loop, 2023.

\bibitem[Zhang et~al.(2021)Zhang, Liniger, Dai, Yu, and
  Gool]{zhang2021endtoend}
Zhejun Zhang, Alexander Liniger, Dengxin Dai, Fisher Yu, and Luc~Van Gool.
\newblock End-to-end urban driving by imitating a reinforcement learning coach,
  2021.

\end{thebibliography}
    
}


\clearpage
\setcounter{page}{1}
\maketitlesupplementary

\section{Roads observations}
\begin{figure}[htb]
		\centering
		\begin{subfigure}[t]{0.42\textwidth}
			\centering
			\includegraphics[trim={0.0cm 1.0cm 0.0cm 2.0cm},clip,width=0.8\textwidth]{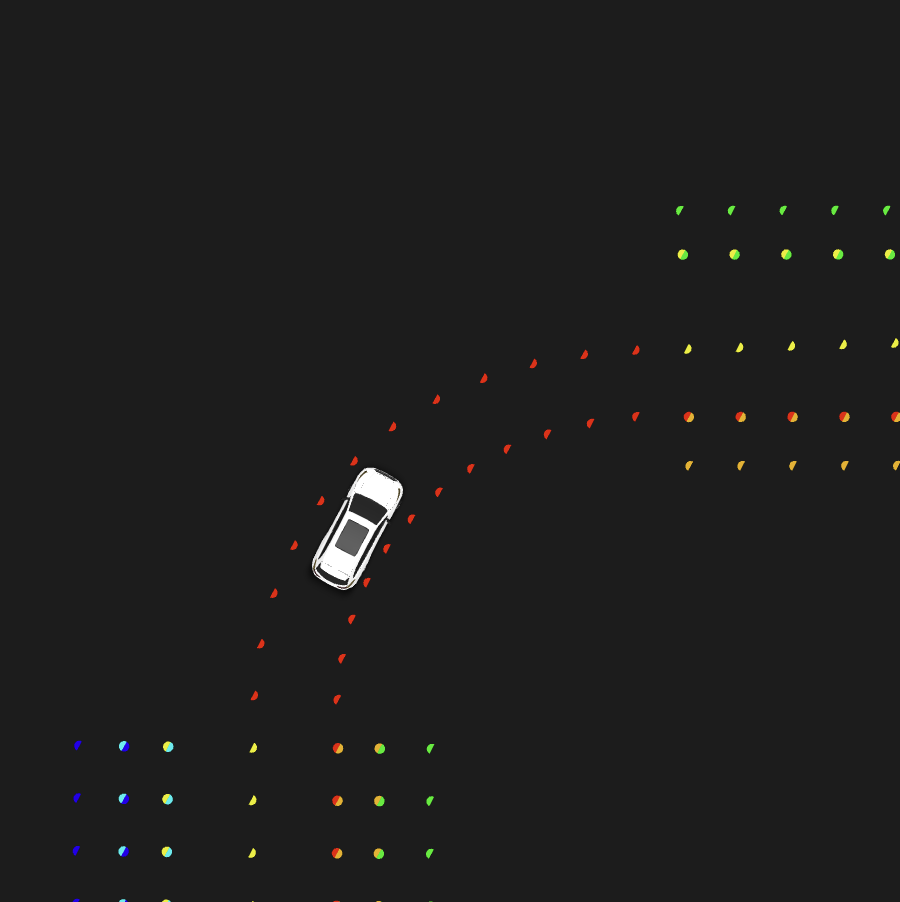}
			\caption{Route border points of all drivable lanes. Only one lane is a valid connection for the right turn the agent is performing. Before and after the turn, other lanes are also valid.}
			\label{fig:sim_lanes}
		\end{subfigure}
		\quad
		\begin{subfigure}[t]{0.42\textwidth}
			\centering
			\includegraphics[trim={0.0cm 1.0cm 0.0cm 2.0cm},clip,width=0.8\textwidth]{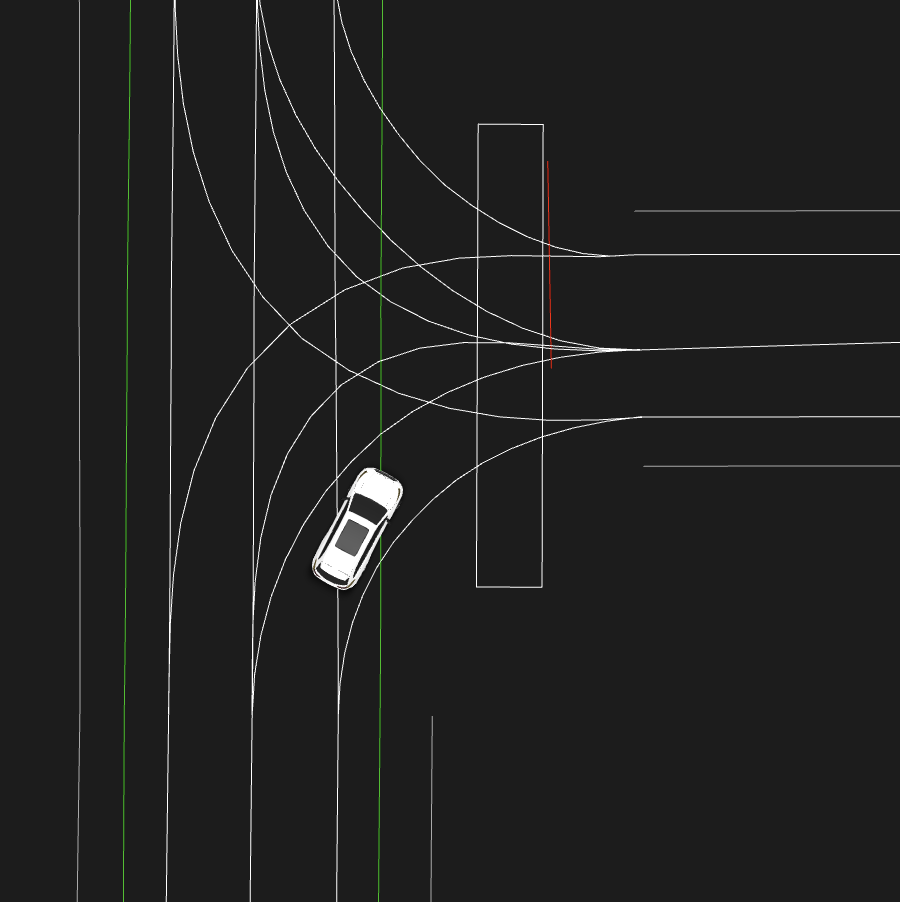}
			\caption{Road network points visualized by poly-lines. The color depends on the properties of the annotation, for example the stop line being visualized in red and bike lane boundaries in green.}
			\label{fig:sim_road_network}
		\end{subfigure}
		\caption{Route and road network observations obtained from the roads library.}
		\label{fig:roads}
	\end{figure}

The implemented roads library in our simulator delivers important information for our rewards and observations. For example, it can calculate the distance to the next stop line. We also use it to create observations of the route and the nearby road segments. This is illustrated in \autoref{fig:roads}.
	
\section{Metric validation}
\label{sec:metric_valitation}
We validated the collision and off-route detection by taking 25 positive and 25 negative samples for each metric from the validation dataset. These have been inspected by human triagers to confirm whether the metric is correct. For the collision and the off-route detection this validation found that all 50 scenarios for each metric were labeled correctly according to our definition. However, overall both metrics were found to be conservative, leading potentially to higher failure rates.

\begin{figure*}[h]
	\centering
	\begin{subfigure}[T]{0.30\textwidth}
		\centering
		\includegraphics[trim={0.5cm 0.5cm 1cm 2cm},clip,width=\textwidth]{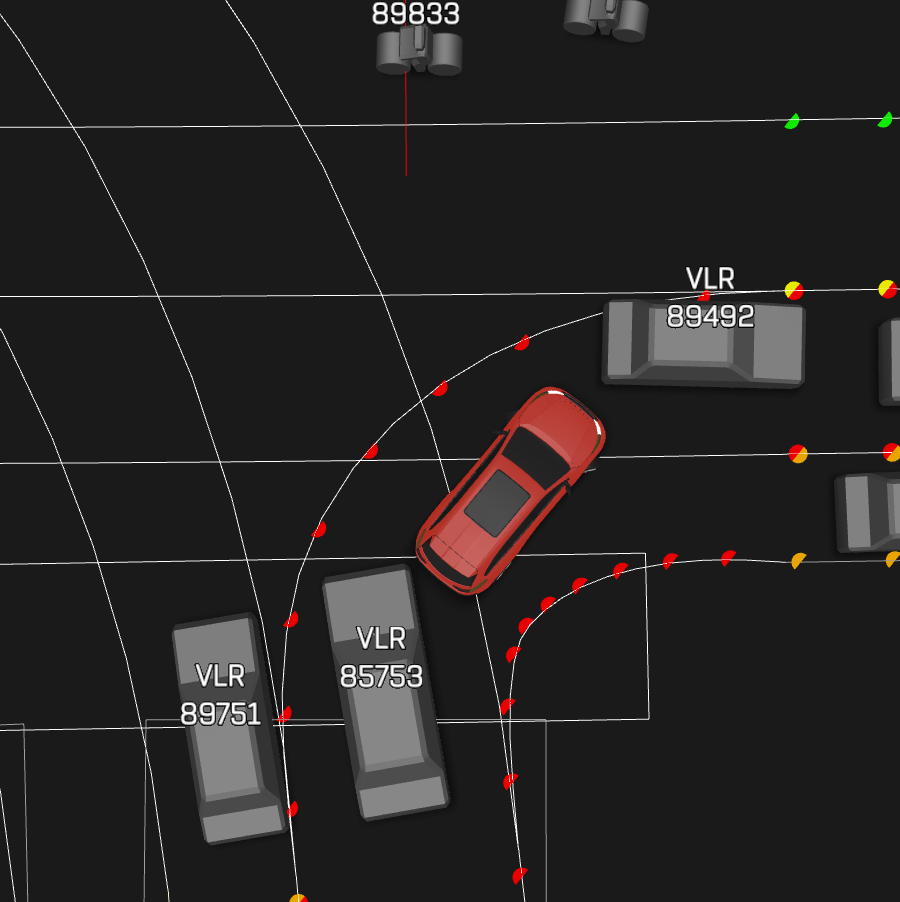}
		\caption{Detected collision with agent 85753 because of the conservative bounding box check.}
		\label{fig:bbox_check}
	\end{subfigure}
	\quad
	\begin{subfigure}[T]{0.30\textwidth}
		\centering
		\includegraphics[trim={0.5cm 0.5cm 1cm 2cm},clip, width=\textwidth]{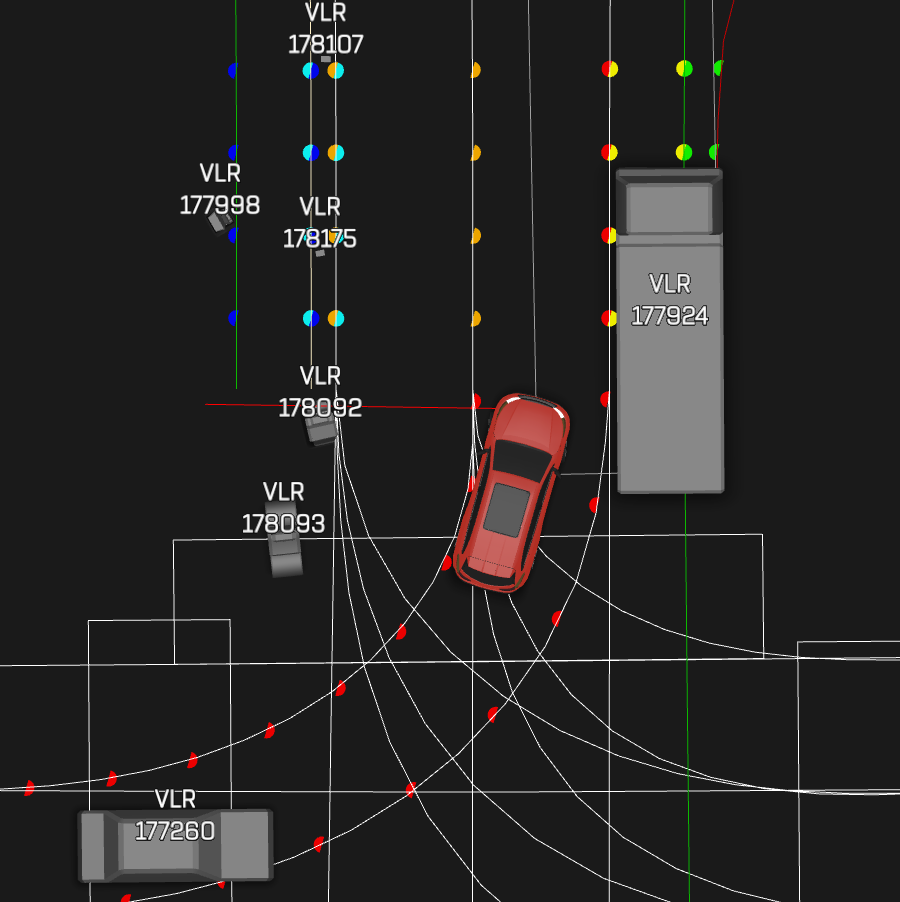}
		\caption{Overly restrictive lane permissibility in a junction. Off-route event detected due to increased bounding box size.}
		\label{fig:off_route_tight}
	\end{subfigure}
	\quad
	\begin{subfigure}[T]{0.30\textwidth}
		\centering
		\includegraphics[trim={0.5cm 0.5cm 1cm 2cm},clip,width=\textwidth]{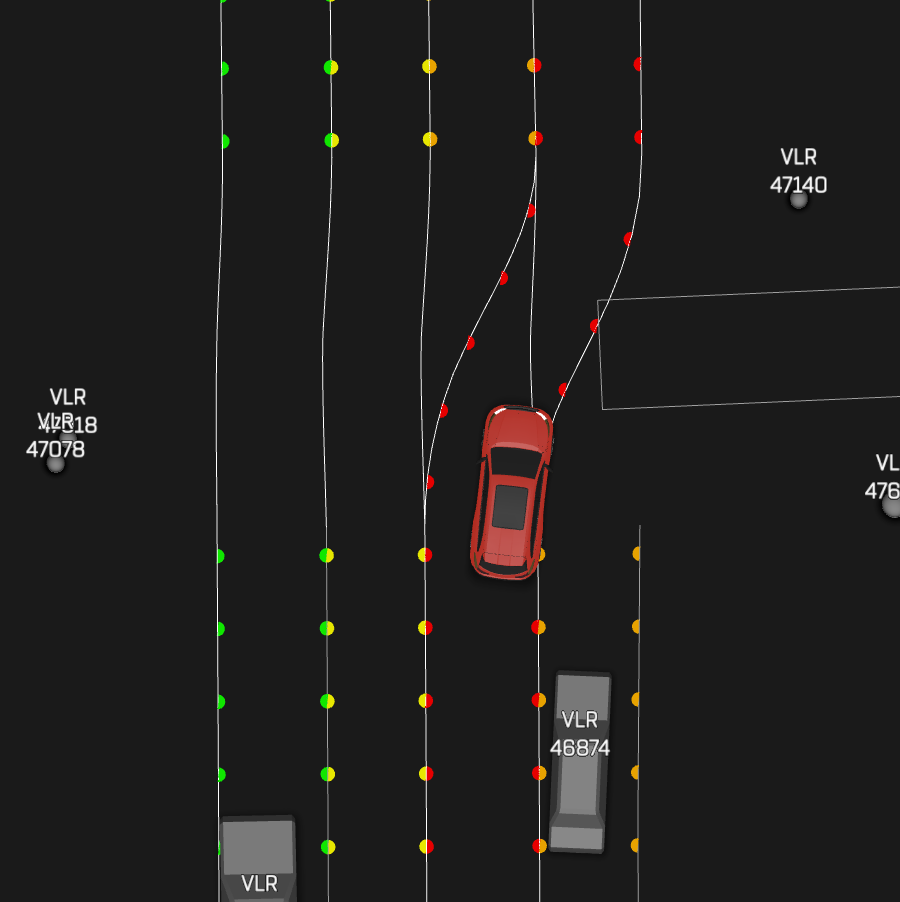}
		\caption{Overly restrictive lane permissibility on a road segment with branching lanes.}
		\label{fig:off_route_lane}
	\end{subfigure}
	\caption{Examples of conservative detection of collisions and off-route events.}
	\label{fig:conservative off}
\end{figure*}

For the collision-free metric a bounding box approximating the vehicle is checked against the bounding box of other vehicles. As the bounding box includes all parts of the vehicle, for example the mirrors and sensors, checking collisions against the bounding box is conservative. \autoref{fig:bbox_check} illustrates this conservative check.

In many situations the permissible lane along the route is overly restrictive leading to off-route events. This occurs particularly in junctions (\autoref{fig:off_route_tight}) and lane merges or branches (\autoref{fig:off_route_lane}). Human expert drivers were found to not exactly follow these lane geometries either. Future work to relax the conditions in these cases to the entire drivable surface is required. On top of that, the check is also on the conservative side due to the increased bounding box size.

\section{Framework scalability}
\label{app:framework_scalability}
When scaling experiments to more GPUs it is important that the policy performance is not affected and that the overall runtime is reduced. Ideally, the reduction in runtime is inversely proportional to the number of GPUs used, so the overall GPU time and, therefore, the cost to run the experiment stays the same.  \autoref{tab:scalability} shows the results for experiments running on 8, 16, and 32 machines. The overall policy performance is very similar and the differences can be considered noise. The overall GPU time sees a marginal uptick. We calculated the normalized GPU time by dividing the total GPU time by the total GPU time of the 8-GPU experiment. As the numbers are close to 1, the framework scales almost perfectly.
\begin{table}[h]
  \caption{Runtime and policy performance metrics for experiments running on different numbers of GPUs. The policy performance is very similar, but the runtime goes down as expected.}
  \label{tab:scalability}
  \centering
  \begin{tabular}[H]{l|lll}
      Number of GPUs & 8 & 16 & 32 \\
      \midrule
      Runtime [h]& $41.58$ & $21.94$ & $11.99$\\
      Total GPU time [h] & $332.6$ & $351.0$ & $383.7$ \\
      Normalized GPU time & $1$  & $1.05$ & $1.15$ \\
      Failure Rate &  $1.28$ & $1.25$ & $1.45$ \\
  \end{tabular}
\end{table}

\section{Ablation of SL pre-training }
\label{app:no_bc}
For this ablation we removed the SL pre-training in order to understand the effectiveness of this step. We used the large 6000 h dataset, the 2.5M model and again trained on 2.5B agent steps.

\begin{figure*}[h]
		\centering
		\begin{subfigure}[T]{0.42\textwidth}
			\centering
			\includegraphics[trim={0.5cm 0.5cm 1cm 2cm},clip,width=\textwidth]{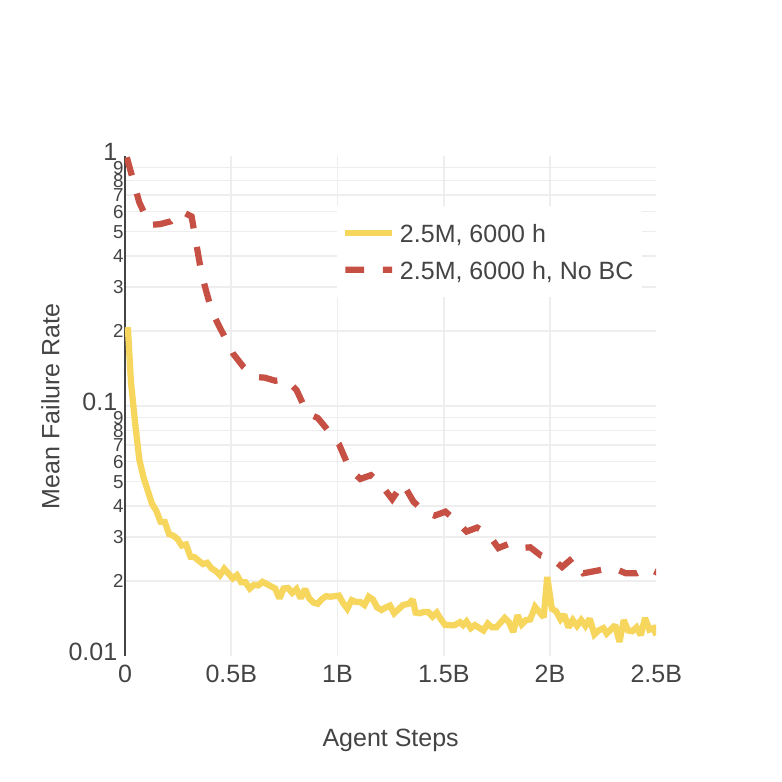}
			\caption{Failure Rate}
			\label{fig:no_bc_failure_rate}
		\end{subfigure}
		\begin{subfigure}[T]{0.42\textwidth}
			\centering
			\includegraphics[trim={0.5cm 0.5cm 1cm 2cm},clip, width=\textwidth]{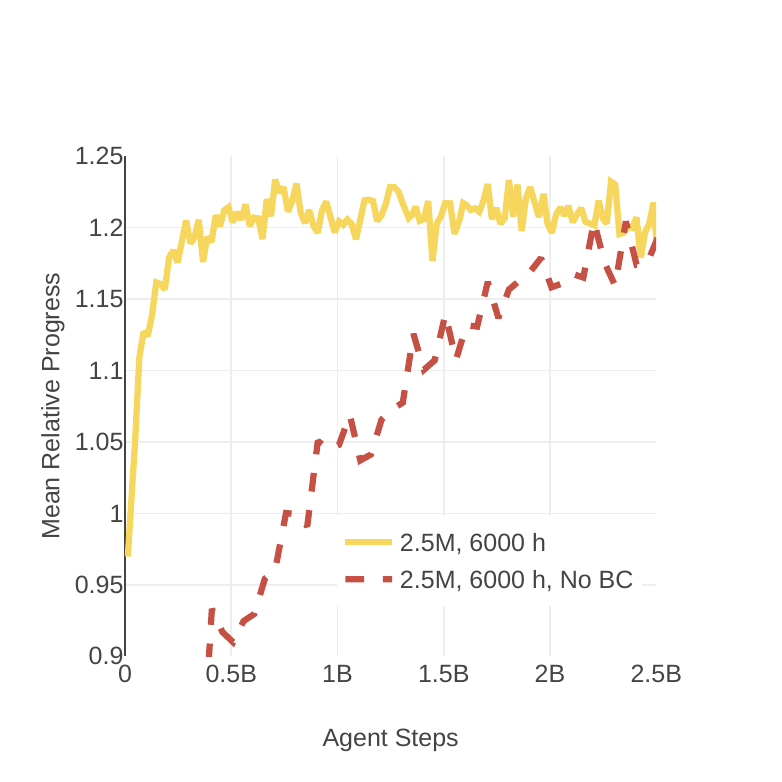}
			\caption{Progress Ratio}
			\label{fig:no_bc_progress_ratio}
		\end{subfigure}
		\caption{Training curves for experiments with and without the SL pre-training on the 2.5M parameter model and the \mbox{6000 h} dataset.}
		\label{fig:no_bc_training_curves}
	\end{figure*}

Even without the SL pre-training the agent can learn to make progress and reduce the failure rate over the course of training as depicted in \autoref{fig:no_bc_training_curves}. However, after 2.5B agent steps the policy without pre-training is still at about 2 \% failure rate, which the pre-trained policy reached already after 0.5B steps. Also the failure rate at 2.5B agent steps is lower for the policy that has been pre-trained. The increased training speed is also observed for the progress ratio. The pre-trained policy reaches values around 120 \% after 0.5B steps and the policy without pre-training catches up to that value at around 2.5B steps. These results overall confirm the effectiveness of the pre-training step in terms of speed. Pre-training also helps to reach better final performance in terms of safety.

\section{Hyperparameters}
\label{app:hyperparameters}
\autoref{tab:bc_hyperparameters} shows the most important hyperparameters for both the behavioral cloning stage and the reinforcement learning stage.
\begin{table}[h]
  \caption{BC and RL Hyperparameters}
  \label{tab:bc_hyperparameters}
  \centering
  \begin{tabular}[H]{l|ll}
      & BC & RL \\
      \midrule
      Batch size & $32768$ & $512$  \\
      Sequence length & - & 32 \\
      Learning rate & $2*10^{-3} $ & $5.6*10^{-5} $ \\
      PPO clip param & -& $0.3$  \\
      Value loss scaling & $10^{-4} $ & $10^{-2} $  \\
      PPO entropy coefficient & - & $3*10^{-2} $  \\
  \end{tabular}
\end{table}

\end{document}